\newcommand{\CLASSINPUTtoptextmargin}{0.75in}
\newcommand{\CLASSINPUTbottomtextmargin}{1in}
\documentclass[conference]{IEEEtran}
\IEEEoverridecommandlockouts
\usepackage{amsmath,amssymb,amsfonts}
\usepackage{algorithmic}
\usepackage{graphicx}
\usepackage{textcomp}
\def\BibTeX{{\rmB\kern-.05em \kern-.05em {\sc i\kern-.025em b}\kern-.08em
    T\kern-.1667em\lower.7ex\hbox{E}\kern-.125emX}}

\usepackage[hyperfootnotes=false]{hyperref}
\usepackage[capitalise]{cleveref}
\usepackage{multirow}
\usepackage[table,xcdraw]{xcolor}
\usepackage{booktabs}
\usepackage{csquotes}

\usepackage[caption=false]{subfig}
\newcommand{\etal}{\textit{et al. }}
\newcommand\blfootnote[1]{%
  \begingroup
  \renewcommand\thefootnote{}\footnote{#1}%
  \addtocounter{footnote}{-1}%
  \endgroup
}
    
\begin{document}

\title{\bfseries Effects of Architectures on Continual Semantic Segmentation }

\author{Tobias Kalb$^{*1}$, Niket Ahuja$^{*1}$, Jingxing Zhou$^{1}$ and J\"urgen Beyerer$^{2, 3}$% <-this % stops a space
\thanks{$^{2}$Tobias Kalb, Niket Ahuja and Jingxing Zhou are
	with Fraunhofer IOSB, 76131 Karlsruhe, Germany {\tt\small \{tobias.kalb,
	niket.ahuja,jingxing.zhou\}@porsche-engineering.de}}%
\thanks{$^{2}$Juergen Beyerer is
	with Fraunhofer IOSB, 76131 Karlsruhe, Germany} %
\thanks{$^{3}$Juergen Beyerer is also with the Vision and Fusion Lab, Karlsruhe Institute
	of Technology KIT, c/o Technologiefabrik, Haid-und-Neu-Strasse 7,
	76131 Karlsruhe, Germany}
}

\maketitle

\begin{abstract}
Research in the field of Continual Semantic Segmentation is mainly investigating novel learning algorithms to overcome catastrophic forgetting of neural networks. 
Most recent publications have focused on improving learning algorithms without distinguishing effects caused by the choice of neural architecture.
Therefore, we study how the choice of neural network architecture affects catastrophic forgetting in class- and domain-incremental semantic segmentation. 
Specifically, we compare the well-researched CNNs to recently proposed Transformers and Hybrid architectures, as well as the impact of the choice of novel normalization layers and different decoder heads.
We find that traditional CNNs like ResNet have high plasticity but low stability, while transformer architectures are much more stable. 
When the inductive biases of CNN architectures are combined with transformers in hybrid architectures, it leads to higher plasticity and stability. 
The stability of these models can be explained by their ability to learn general features that are robust against distribution shifts.
Experiments with different normalization layers show that Continual Normalization achieves the best trade-off in terms of adaptability and stability of the model.
In the class-incremental setting, the choice of the normalization layer has much less impact.
Our experiments suggest that the right choice of architecture can significantly reduce forgetting even with naive fine-tuning and confirm that for real-world applications, the architecture is an important factor in designing a continual learning model.
\end{abstract}


\blfootnote{\textsuperscript{*} Both authors contributed equally to this work.}

\section{INTRODUCTION}
With the rise of large-scale datasets and deep-learning architectures for Computer Vision, semantic segmentation is now widely used for environment perception in automated vehicles, in which it is entrusted to reliably recognize and comprehend pictures at the pixel level.
To deploy such semantic segmentation models at a large scale for highly automated driving across multiple countries and environments, they are not only required to be adapted to each new environment but also need to be constantly updated to cope with new and different-looking objects (e.g. e-scooter), unforeseen prediction errors. 
However, when a neural network is updated incrementally without training on the entire previously collected dataset it suffers from a severe drop in performance on previous tasks.
This effect is referred to as catastrophic forgetting~\cite{MCCLOSKEY1989109}. 
On-going research in continual semantic segmentation is focused on developing new learning algorithms to mitigate the effects of catastrophic forgetting, while always using similar convolutional neural network (CNN) architectures. 
Recent work pointed out that architecture choices such as width, depth and normalization layers can be important ingredients to combat the effects of catastrophic forgetting~\cite{Architecturematters}. 
Therefore, in this work we aim to study how the recent development in computer vision models such as Vision Transformers (VTs), improved CNNs, hybrid models and new normalization layers affect catastrophic forgetting in Continual Semantic Segmentation (CSS).
We find that traditional CNNs like ResNet have high plasticity but low stability, whereas VT architectures are much more stable. When the inductive biases of CNN architectures are combined with VTs in hybrid architectures, we find a good trade-off between plasticity and stability. The stability of hybrid and transformer architectures can be explained by their ability to learn general features that are robust against distribution shifts. Our findings suggest that architectural choices are crucial for the development of continual learning methods.
\begin{enumerate}
    \item The increased robustness towards forgetting of transformers is not an inherent feature of the self-attention introduced by the transformer, but potentially by other micro and macro design choices, as we confirm in our ablation study on ConvNeXt \cite{Convnext}.
    \item The choice of normalization layers is especially important in the domain-incremental setting, where Continual Normalization has the best trade-off in terms of plasticity and stability.
    \item While in class-incremental learning forgetting mainly occurs in the decoder, we find that the backbone architecture still plays an important role in reducing forgetting and that input from multiple-stages of the encoder into the decoder can have a detrimental effect on forgetting, as activations in early stages of the network are more easily affected.
\end{enumerate}

\section{Related Works}
\subsection{Continual Semantic Segmentation}
The majority of research in continual semantic segmentation can be broadly divided into class-incremental semantic segmentation (CiSS) and continual unsupervised domain adaptation (CUDA). 
The challenges of CiSS are mostly addressed by using a knowledge distillation-based loss with hard or soft pseudo labels \cite{Michieli2019,klingner2020class,Cermelli_2022_CVPR,Douillard2020}, addressing the background shift \cite{Cermelli2020}, disentangling the representation of individual classes \cite{Michieli2021} or combining knowledge-distillation with replay \cite{Kalb_itsc,Kalb2021,Douillard2021,Maracani_2021_ICCV}. 
The goal of CUDA is to adapt a model trained on a supervised source dataset to a sequence of unlabeled domains. 
To compensate for the missing labels of the target domains, the model has access to the original source dataset throughout all training phases. 
Most methods in this category store information about a specific domain's style in order to transform source images into the styles of the target domains during training. 
Recent work achieves this by storing low-frequency components of the images for every domain \cite{9564566}, by capturing the style of the domain with generative models \cite{Wu_2019_ICCV, cuda_robert} or by using a domain-specific memory to mitigate forgetting \cite{cuda2}. 
Supervised domain-incremental learning is rarely considered \cite{Kalb2021}. 
Because our primary goal is to evaluate architectures in CSS, we evaluate methods without regard to any specific continual learning algorithm. 

\subsection{Transformers vs. CNNs}\label{sec:transformersarerobust}
With the well-established CNNs and the rise of the Vision Transformers (VT), there are currently two competing architecture types in the field of computer vision, each with very distinct properties. Previous work has extensively studied how the performance of these architectures differs on various computer vision tasks. Recent work claims that VTs are more robust than CNNs, specifically to severe domain shift~\cite{robtrans2,robtrans4}, adversarial attacks~\cite{robtrans1,robtrans3} or perturbations~\cite{robtrans5}.
Some of them also indicate that the comparisons in other papers have been distorted because of different scales and distinct training frameworks~\cite{robCNN1} and claim that CNNs can easily match the robustness of VTs~\cite{robCNN2}. 
More recently, Bai \etal \cite{robCNN1} indicate that prior comparisons of CNNs and VTs have been unfair due to the unique scales and training schemes used for transformers~\cite{robCNN1}. They show that CNNs are just as robust as VTs when comparable training schemes are used and that robustness is not an inherent feature of the attention-mechanism~\cite{robCNN1}. 
Another interesting finding is that CNNs tend to have a texture bias, whereas VTs have a shape bias~\cite{cnntranbias1,robtrans1}.

\subsection{Architectures in Continual Learning}
Previous work on architectures for continual learning focused on optimizing the architecture for continual learning using neural architecture search that focuses on how to efficiently share or expand the model for continual learning ~\cite{Param3,NAS2,NAS3}.
To the best of our knowledge, only Mirzadeh \etal~\cite{Architecturematters} investigate the role of architectures in continual learning, but they only look at nominal recognition tasks rather than semantic segmentation tasks. Furthermore, they conduct a study only on simply stacked CNN layers rather than a backbone architecture. While they also study VT for continual learning, they do not consider other transformer or hybrid architectures.
A recent report from the ICCV 2021 Challenge SSLAD-Track3B suggests that that the recent Swin Transformer suffers less from forgetting than its CNN counterparts and thus performs better in continual learning~\cite{transformer_better_CL}.

\section{Problem Formulation}
The goal in semantic segmentation is to assign one of the predefined classes $\mathcal{C}$ to each pixel in a given image. Therefore, given a task $T$ the model $f$ has to learn a mapping $f: \mathcal{X} \to \mathbb{R}^{H\times W\times |\mathcal C|}$ from the image space $\mathcal{X}$ to a score vector $\hat{y} \in  \mathbb{R}^{H\times W\times |\mathcal C|}$. 
Typically the training task $T = \{(x_n, y_n)\}^{N}_{n=1}$ consists of a set of $N$ images $x_n\in \mathcal{X}$ with $\mathcal{X} = \mathbb{R}^{H \times W \times 3}$ and corresponding labels $y_n\in \mathcal{Y}$ with $\mathcal{Y} = \mathcal{C}^{H \times W}$. 
In continual learning $f$ is incrementally optimized on a sequence of tasks $T_k$, which in this work either introduce a new set of classes or a change in the input distribution. In class-incremental learning each task $T_k$ extends the previous set of classes $\mathcal{C}_{k-1}$ by a set of novel classes $\mathcal{S}_k$ resulting in the new label set $\mathcal{C}_k = \mathcal{C}_{k-1} \cup \mathcal{S}_k$. 
In contrast, in domain-incremental learning the classes stay the same, but the images are obtained from the different input distributions, and therefore have distinct visual appearance.
Catastrophic forgetting occurs when $f$ is optimized on the data of task $T_k$ and disregards the data from previous tasks $T_{i}$ with $i<k$.
In this work we aim to empirically study which architectures for the model $f$ perform the best in continual semantic segmentation. 

\section{Experiments}
In this study, we aim to examine how different neural network architectures and normalization layers perform in continual semantic segmentation. 
We conduct these experiments for both class- and domain-incremental learning because it has been observed that catastrophic forgetting affects neural networks differently in these scenarios~\cite{Kalb2022,Kalb2021}.

\begin{figure}
	\centering
 \fontsize{9pt}{11pt}
    \def\svgwidth{1.001\columnwidth}
	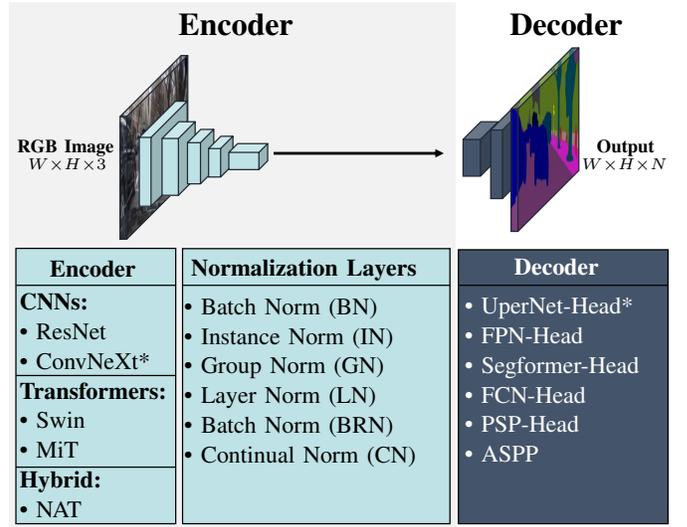
	\caption{Experimental Setup and architecture choices. In all our experiments, we use the same UperNet Decoder head and only exchange the Backbone with different VT, CNN and hybrid architectures. We investigate the impact of the Decoder using a ConvNeXt backbone.}
	\label{fig:experiment_overview}
\end{figure}

\subsection{Experimental Setup}\label{chap:experimental setup}
\paragraph*{Architectures} In our fist experiments, we aim to evaluate the effect of different backbone architectures for continual semantic segmentation, so we only exchange the encoder of the semantic segmentation network and use the same UperNet Decoder Head~\cite{Upenet}. 
We select these backbones from different neural network architecture types like CNNs, VTs and hybrid architectures to compare the effect of the inherent architectural differences between them. To achieve a fair comparison between the different architectures, we choose the backbone architectures such that they have a similar number of parameters, compare \cref{tab:arch_overview}. From CNNs, we use the well-established ResNet-50~\cite{resnet50} and the state-of-the-art ConvNext-T~\cite{Convnext} that modernizes the ResNet architecture with design decisions used in VTs. As representative for the VTs we select the Swin Transformer, as previous studies suggested its superior performance in continual learning to its CNN counterparts~\cite{transformer_better_CL}. Additionally, we use MiT, because contrary to the other backbones, it was specifically designed for segmentation tasks. 
Finally, we also evaluate the neighborhood attention transformer (NAT)\cite{hassani2022neighborhood} as a backbone architecture, as it effectively introduces inductive biases similar to CNN using overlapping windows and applies attention within the window using a novel attention mechanism called neighborhood attention.
The additional inductive bias should make them less data-dependent than their VT counterparts. All the selected architectures are summarized in \cref{tab:arch_overview} along with the type of architecture and number of parameters in the network. 
Furthermore, we also evaluate different normalization layers, as recent work has shown that they also significantly contribute to catastrophic forgetting~\cite{Lomonaco_2020_CVPR_Workshops,bnforget}. 
Specifically, we exchange the BatchNorm Layers of ResNet-50 with Continual- \cite{CN}, Group- \cite{wu2018group}, Instance- \cite{ulyanov2016instance}, Layer and Batch Re-Normalization layers \cite{ioffe2017batch} and evaluate the resulting networks in the two incremental scenarios.

\begin{table}
\caption{Overview of the selected Backbone architectures with their respective Top-1 ImageNet accuracy}
\label{tab:arch_overview}
\centering
\resizebox{0.8\columnwidth}{!}{%
\begin{tabular}{@{}llccc@{}}
\toprule
\textbf{Type} & \textbf{Name} & \textbf{\#params} & \textbf{FLOPs} & \textbf{Top-1} \\ \midrule
 & \textbf{ResNet-50} & 23M & 4.1G & 78.8 \\
\multirow{-2}{*}{\textbf{CNN}} & \textbf{ConvNeXt-T} & 29M & 4.5G & 82.1 \\ \midrule
 & \textbf{Swin-T} & 28M & 4.5G & 81.3 \\
\multirow{-2}{*}{\textbf{Transformer}} & \textbf{MiT-b2} & 25.2M & 4.0G & 81.6 \\ \midrule
\textbf{Hybrid} & \textbf{NAT-T} & 30M & 4.3G & 83.2 \\ \bottomrule
\end{tabular}%
}
\end{table}

\paragraph*{Datasets} We evaluate the architectures in both a class- and domain-incremental setting. For the class-incremental setting we utilize PascalVoC-2012 dataset with most commonly used 15--5 split, which means that we learn the classes 0--15 in the first task and the classes 16--20 in the second task~\cite{Cermelli2020}. We choose the \textit{Overlapped} setting, in which classes from future task can occur in the current task, though labeled as background.
In our domain-incremental experiments, we adapt to adverse weather conditions using an incremental Cityscapes (CS) to ACDC setup.
CS is a dataset collected during daytime in dry weather conditions, across German, Swiss, and French cities.
In contrast, ACDC was recorded during different adverse weather conditions and divided into four different subsets: \textit{Night}, \textit{Rain}, \textit{Fog} and \textit{Snow}. 
The increment in ACDC and CS is purely domain-incremental since they share a common 19-class labelling policy.

\paragraph*{Optimization Strategy} We train the networks using AdamW as optimizer with a base learning rate of $1 \times 10^{-4}$, which we tune for the different backbone architectures. 
We begin training with a 1500-iteration warm-up phase, followed by a linear decaying learning rate schedule. We use a batch size of 6 and use the same learning rate and schedule for the second task.
In the domain-incremental setting, we train the models for 120 epochs and use a input resolution of $769 \times 769$. On Pascal-15--5, due to the bigger dataset size, we train the models only for 40 epochs with a input resolution of $512 \times 512$. 
For evaluation we use the individual validation sets without any scaling or cropping. As previous work has noted that in the class-incremental learning setting simply fine-tuning the model to new data leads to a severe background bias~\cite{Kalb2022}, we train the models with the unbiased-cross-entropy loss (UNCE)~\cite{Cermelli2020}, which models the uncertainty of the background class.

\paragraph*{Evaluation Metrics} We evaluate every model on the validation set of each dataset using its mean intersection-over-union (mIoU). 
We denote the mIoU of a model trained until task $p$ and evaluated on $q$ is $\text{mIoU}_{p, q}$. 
For instance, a model trained until $p=1$ and tested on $q=0$ would be denoted by $\text{mIoU}_{1, 0}$. 
Moreover, we report \textit{average learning accuracy} and \textit{forgetting} that measure the learning capability and the severity of forgetting~\cite{Architecturematters}. 
\begin{equation}
    \text{average learning acc.} = \frac{\text{mIoU}_{0, 0} + \text{mIoU}_{1, 1}}{2}
\end{equation}
\begin{equation}
    \text{forgetting} = \text{mIoU}_{0, 0} - \text{mIoU}_{1, 0}
\end{equation}

\subsection{Domain-Incremental Learning} \label{di_results}
We report the results for the different selected backbone architectures in \cref{tab:result_di_arch} and for different normalization layers in \cref{tab:result_di_norm}. 
We observe that the CNN backbones show a high learning accuracy that is competitive with recent ViTs, but in the case of ResNet-50 are more affected by forgetting compared to the remaining backbones. 
The transformer backbones MiT-b2 and Swin-T significantly mitigate the effects of forgetting, but in turn suffer a moderate drop in learning accuracy. 
This could potentially be explained by the fact that they are known to be more data-dependent than their CNN counterparts~\cite{dosovitskiy2020image}. 
The hybrid model NAT-T shows good performance for both learning accuracy and forgetting, as it re-introduces the inductive bias found in Convolution Layers, and therefore is not required to learn those biases in the early layers \cite{ raghu2021vision,hassani2022neighborhood}.
Furthermore, we see that in the domain-incremental setting ConvNeXt-T achieves the overall best results, indicating that the improved robustness towards forgetting of the transformer backbones is not an inherent feature of the self-attention mechanisms, but can be attributed to other architectural changes, which we will discuss in \cref{sec:convnext}. Finally, we note that the ConvNext-T and the transformer-based backbones also achieve higher mIoU on ACDC after only training on Cityscapes, which indicates that the reduction of forgetting may partly be caused by better generalization of these models.
Next, we compare the performance of the different normalization layers for domain-incremental learning in \cref{tab:result_di_norm}. 
The results show that the standard Batch Norm Layer of ResNet-50 is most affected by forgetting, while also being the best at adapting to the new domains. 
We hypothesize this is due to a severe bias of the changing population mean and variance of the Batch Norm Layer to the most recent task.
Batch Re-Norm alleviates the discrepancy between the changing population statistics, but still reduces forgetting only very slightly. 
Channel normalization layers such as Group- and Instance-Norm are most effective in mitigating forgetting in the domain-incremental setting, but at the same time lead to reduced learning accuracy. 
Layer Normalization, which is also used in the selected transformer backbones, also improves performance on previous tasks, but degrades performance on new tasks. 
As a whole, Continual Norm is the most effective since it matches the learning accuracy of BN while is also significantly reduces forgetting.
This is because CN combines the benefits of Batch and Group Norm by first normalizing along the channel dimension and then normalizing along the mini-batch dimension, thereby achieving both flexibility and stability.

\begin{table}
\caption{Results in the Domain-Incremental Setting with different Backbone Architectures.}
\label{tab:result_di_arch}
\resizebox{\columnwidth}{!}{%
\begin{tabular}{@{}l|cc|cc|cc}
\toprule
\multicolumn{1}{c|}{} & \multicolumn{2}{c|}{\textbf{Task 1 (CS) }} & \multicolumn{2}{c|}{\textbf{Task 2 (ACDC)}} & \multicolumn{1}{l}{\textit{learning}} & \multicolumn{1}{l}{} \\
\multicolumn{1}{c|}{\multirow{-2}{*}{\textbf{Method}}} & \textit{Cityscapes} & \textit{ACDC} & \textit{Cityscapes} & \textit{ACDC} & \textit{accuracy} & \textit{forgetting} \\ \midrule
ResNet-50 & \cellcolor[HTML]{FA9473}78.6 & \cellcolor[HTML]{F8696B}37.9 & \cellcolor[HTML]{F8696B}60.7 & \cellcolor[HTML]{63BE7B}72.4 & \cellcolor[HTML]{C1DA81}75.5 & \cellcolor[HTML]{F8696B}17.9 \\
ConvNeXt-T & \cellcolor[HTML]{63BE7B}80.9 & \cellcolor[HTML]{63BE7B}51.6 & \cellcolor[HTML]{63BE7B}71.9 & \cellcolor[HTML]{7FC77D}71.9 & \cellcolor[HTML]{63BE7B}76.4 & \cellcolor[HTML]{63BE7B}9 \\
Swin-T & \cellcolor[HTML]{FFEB84}78.8 & \cellcolor[HTML]{FEDC81}44.2 & \cellcolor[HTML]{FDCE7E}64.7 & \cellcolor[HTML]{FFEB84}69.6 & \cellcolor[HTML]{FCC37C}74.2 & \cellcolor[HTML]{FEC87E}14.1 \\
MiT-b2 & \cellcolor[HTML]{F8696B}78.5 & \cellcolor[HTML]{FFEB84}45 & \cellcolor[HTML]{FFEB84}65.8 & \cellcolor[HTML]{F8696B}66.8 & \cellcolor[HTML]{F8696B}72.6 & \cellcolor[HTML]{FFEB84}12.7 \\
NAT-T & \cellcolor[HTML]{89C97E}80.4 & \cellcolor[HTML]{F1E784}45.6 & \cellcolor[HTML]{9ECF7F}69.6 & \cellcolor[HTML]{FEE182}69.4 & \cellcolor[HTML]{FFEB84}74.9 & \cellcolor[HTML]{AED37F}10.8 \\ \bottomrule
\end{tabular}%
}
\end{table}

\begin{table}
\caption{Results in the Domain-Incremental Setting with different Normalization Layers.}
\label{tab:result_di_norm}
\resizebox{\columnwidth}{!}{%

\begin{tabular}{@{}l|cc|cc|cc}
\toprule
\multicolumn{1}{c|}{} & \multicolumn{2}{c|}{\textbf{Task 1 (CS)}} & \multicolumn{2}{c|}{\textbf{Task 2 (ACDC)}} & \textit{learning} &  \\

\multicolumn{1}{c|}{\multirow{-2}{*}{\textbf{Normalization Layer}}} & \textit{Cityscapes} & \multicolumn{1}{c|}{\textit{ACDC}} & \textit{Cityscapes} & \multicolumn{1}{c|}{\textit{ACDC}} & \textit{accuracy} & \textit{forgetting} \\ \midrule
Batch Norm (BN) & \cellcolor[HTML]{76C47D}78.6 & \multicolumn{1}{c|}{\cellcolor[HTML]{FDCE7E}37.9} & \cellcolor[HTML]{FFEB84}60.7 & \multicolumn{1}{c|}{\cellcolor[HTML]{63BE7B}72.4} & \cellcolor[HTML]{63BE7B}75.5 & \cellcolor[HTML]{F8696B}17.9 \\
Continual Norm (CN) & \cellcolor[HTML]{63BE7B}79 & \multicolumn{1}{c|}{\cellcolor[HTML]{65BF7C}48} & \cellcolor[HTML]{63BE7B}64.8 & \multicolumn{1}{c|}{\cellcolor[HTML]{7AC57D}72.1} & \cellcolor[HTML]{63BE7B}75.5 & \cellcolor[HTML]{C1D980}14.2 \\
Batch Re-Norm (BRN) & \cellcolor[HTML]{AFD480}77.4 & \multicolumn{1}{c|}{\cellcolor[HTML]{EDE683}40.1} & \cellcolor[HTML]{FFEB84}60.7 & \multicolumn{1}{c|}{\cellcolor[HTML]{B0D480}71.4} & \cellcolor[HTML]{A8D27F}74.4 & \cellcolor[HTML]{FBA076}16.7 \\
Group Norm (GN) & \cellcolor[HTML]{F8696B}71.5 & \multicolumn{1}{c|}{\cellcolor[HTML]{FDCB7D}37.8} & \cellcolor[HTML]{FA9573}57.6 & \multicolumn{1}{c|}{\cellcolor[HTML]{F8696B}66} & \cellcolor[HTML]{F8696B}68.7 & \cellcolor[HTML]{ABD27F}13.9 \\
Instance Norm (IN) & \cellcolor[HTML]{FCB679}74.0 & \multicolumn{1}{c|}{\cellcolor[HTML]{63BE7B}48.1} & \cellcolor[HTML]{F0E784}61.1 & \multicolumn{1}{c|}{\cellcolor[HTML]{FDCB7D}69.3} & \cellcolor[HTML]{FCC07B}71.6 & \cellcolor[HTML]{63BE7B}12.9 \\
Layer Norm (LN) & \cellcolor[HTML]{F8756D}71.9 & \cellcolor[HTML]{F8696B}34 & \cellcolor[HTML]{F8696B}56 & \cellcolor[HTML]{F8696B}66 & \cellcolor[HTML]{F86F6C}68.9 & \cellcolor[HTML]{FDC57D}15.9 \\ \bottomrule
\end{tabular}%
}
\end{table}

\subsection{Class-Incremental Learning}
\begin{figure*}
    \centering
    \subfloat[ResNet50]{\fontsize{6pt}{7pt} \def\svgwidth{0.22\textwidth} 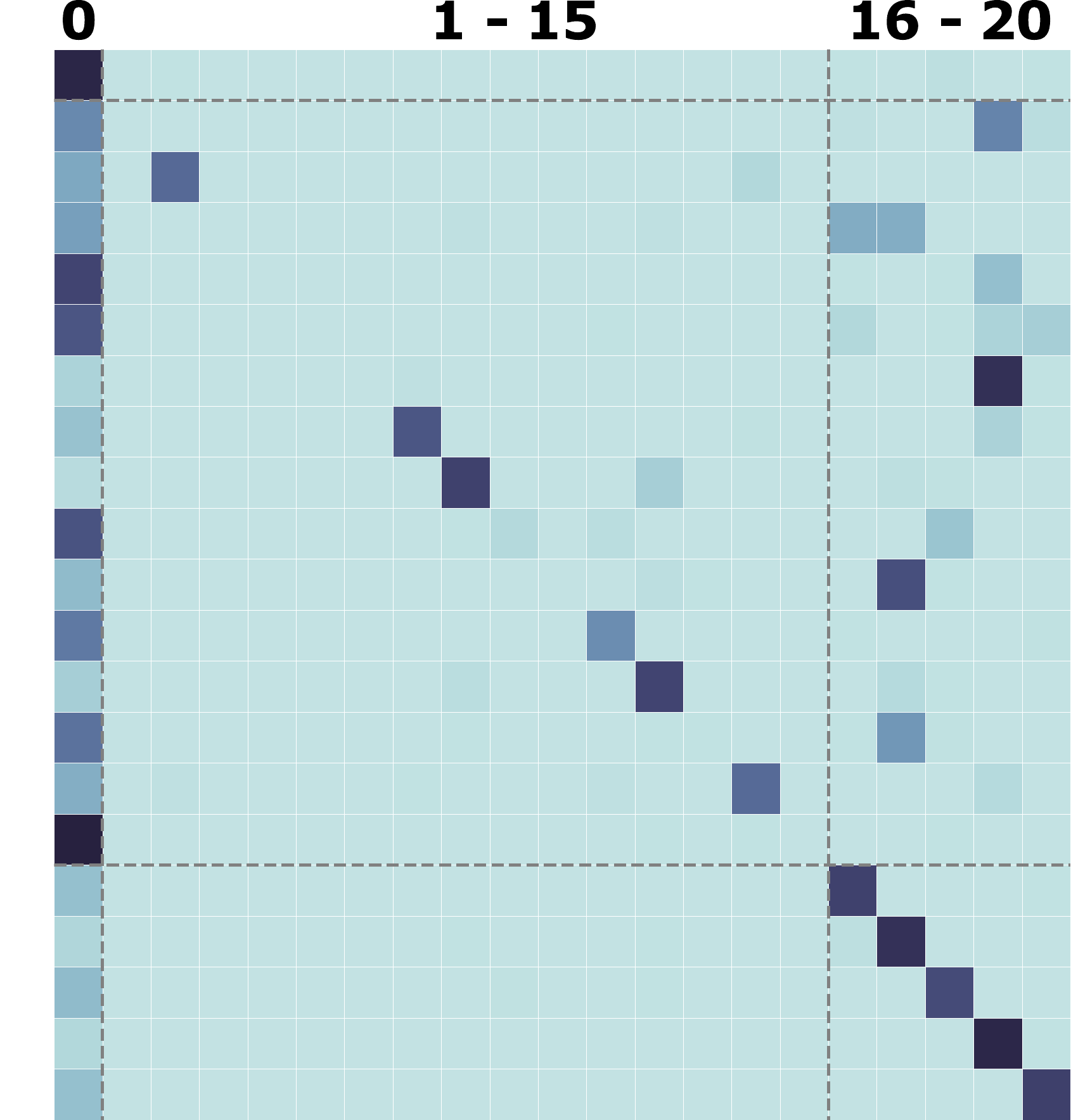}\qquad
    \subfloat[Swin-T]{\fontsize{6pt}{7pt} \def\svgwidth{0.22\textwidth} 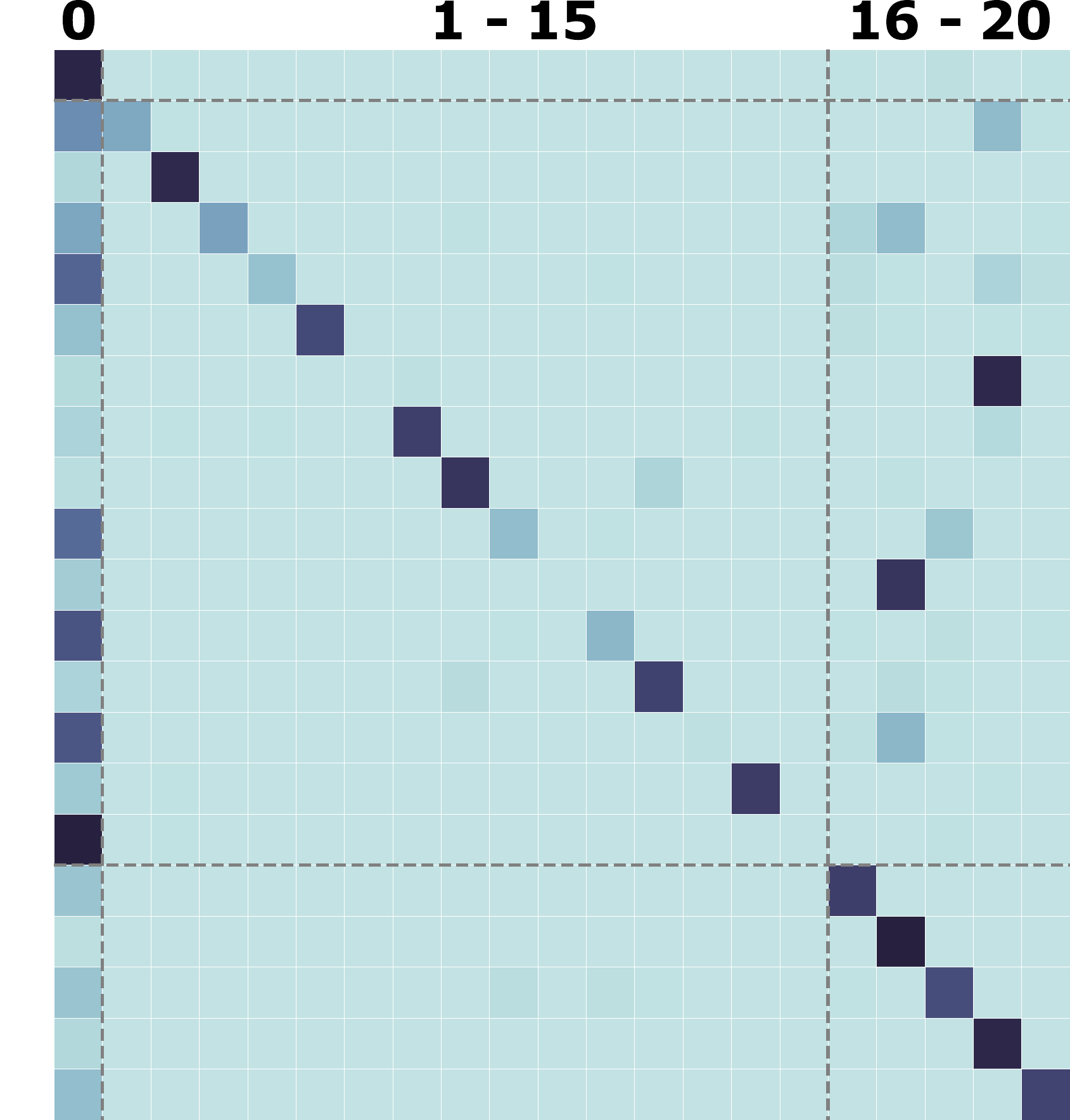}\qquad
    \subfloat[ConvNeXt-T]{\fontsize{6pt}{7pt} \def\svgwidth{0.22\textwidth} 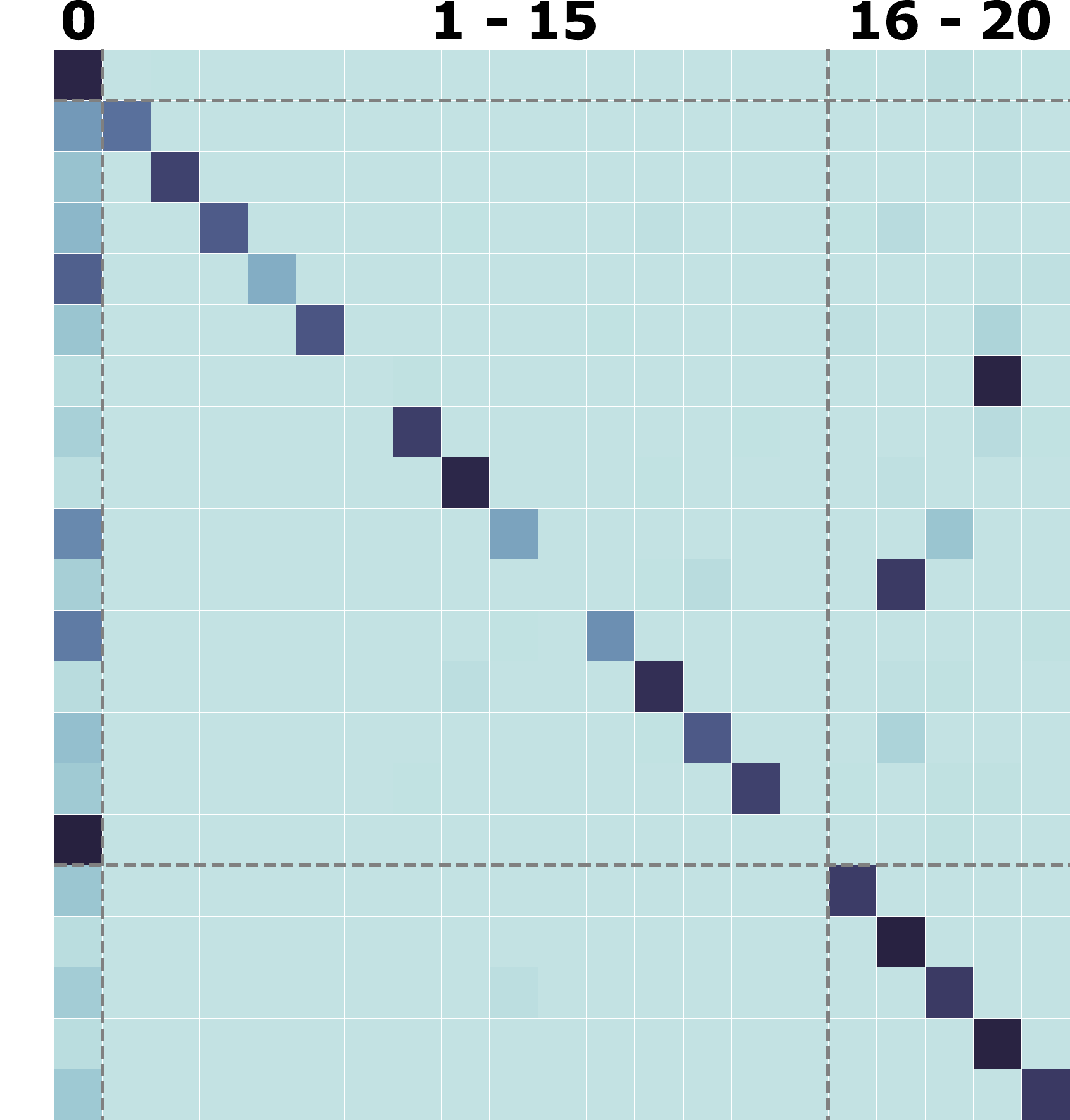}

    \resizebox{\textwidth}{!}{%
    \begin{tabular}{@{}lllllllllllllllllllll@{}}
    \toprule
    \multicolumn{1}{c}{0} & \multicolumn{1}{c}{1} & \multicolumn{1}{c}{2} & \multicolumn{1}{c}{3} & \multicolumn{1}{c}{4} & \multicolumn{1}{c}{5} & \multicolumn{1}{c}{6} & \multicolumn{1}{c}{7} & \multicolumn{1}{c}{8} & \multicolumn{1}{c}{9} & \multicolumn{1}{c}{10} & \multicolumn{1}{c}{11} & \multicolumn{1}{c}{12} & \multicolumn{1}{c}{13} & \multicolumn{1}{c}{14} & \multicolumn{1}{c}{15} & \multicolumn{1}{c}{16} & \multicolumn{1}{c}{17} & \multicolumn{1}{c}{18} & \multicolumn{1}{c}{19} & \multicolumn{1}{c}{20} \\ \midrule
    background & aeroplane & bicycle & bird & boat & bottle & bus & car & cat & chair & cow & dining table & dog & horse & motorbike & person & potted plant & sheep & sofa & train & monitor \\ \bottomrule
    \end{tabular}
    }
    \label{fig:confmat}
    \caption{Confusion matrices after training on PascalVoc-15--5. The confusion matrix for ResNet-50 shows a severe bias to the background class and classes of the recent task (16--20). Using Swin or ConvNeXt as backbone with the same decoder head decreases the bias for the new classes and the background class. }
\end{figure*}

The results for the same architectures for the class-incremental Pascal-15-5 setting are displayed in \cref{tab:result_ci_arch}. 
Overall, we observe that forgetting in the class-incremental setting is much more severe than in the domain-incremental setting.
Similar to the results in \cref{di_results}, the VT and hybrid networks, namely Swin and NAT, are more effective at mitigating forgetting than ResNet-50.
However, MiT is an outlier and surprisingly more affected by forgetting than ResNet-50.
ConvNeXt performs the best in the class-incremental setting with significant improvements compared to ResNet-50, in both learning new classes and mitigating the forgetting of old classes.
Unlike domain-incremental results, it is observed that VT have higher learning accuracy than ResNet in the class-incremental setting.
The confusion matrices in \cref{fig:confmat} show that even with using UNCE, we still see a severe bias for the background class and new classes, especially for ResNet-50.
ConvNeXt and the transformer-based models significantly reduce this bias, which indicates that the features of the VTs and ConvNeXt are more useful for the decoder to discriminate between the different classes.
Furthermore, we note that the only classes that ConvNeXt is not able to learn correctly are \textit{Cow} and \textit{Bus}, which are confused as the new classes \textit{sheep} and \textit{train}, respectively. 
The results show that although forgetting in the class-incremental setting is mainly affecting decoder layers \cite{Kalb2022,davari2021probing}, the choice of the backbone still has a significant impact on reducing forgetting.
When comparing the different normalization layers in class-incremental learning, we observe that Batch Norm, Batch Re-Normn and Continual Norm have higher learning compared to channel normalization layers, which again confirms that normalising along the mini-batch dimension improves the forward transfer capability. 
However, unlike in the domain-incremental setting, Continual Norm does not mitigate forgetting compared to the Batch Norm or methods that normalize across channel dimension. 
We explain this by the fact that in the class-incremental setting the population statistics of the encoder are not changing as much, as the input data distribution is much more similar than in the domain-incremental setting. 

\begin{table}
\caption{Results in the Class-Incremental Setting with different backbone architectures.}
\label{tab:result_ci_arch}
\resizebox{\columnwidth}{!}{%
\begin{tabular}{@{}l|c|ccc|cc}
\toprule
\multicolumn{1}{c|}{} & \textbf{Task 1} & \multicolumn{3}{c|}{\textbf{Task 2}} & \textit{learning} &  \\
 
\multicolumn{1}{c|}{\multirow{-2}{*}{\textbf{Method}}} & \textit{0-15} & \textit{0-15} & \textit{16-20} & \textit{all} & \textit{accuracy} & \textit{forgetting} \\ \midrule
ResNet-50 & \cellcolor[HTML]{F8696B}74.9 & \cellcolor[HTML]{F98A71}23.5 & \cellcolor[HTML]{F8696B}32.4 & \cellcolor[HTML]{F87A6E}25.7 & \cellcolor[HTML]{F8696B}53.7 & \cellcolor[HTML]{FCB37A}51.4 \\
ConvNeXt-T & \cellcolor[HTML]{72C37C}81.1 & \cellcolor[HTML]{63BE7B}50.4 & \cellcolor[HTML]{63BE7B}43.9 & \cellcolor[HTML]{63BE7B}48.9 & \cellcolor[HTML]{63BE7B}62.5 & \cellcolor[HTML]{63BE7B}30.7 \\
Swin-T & \cellcolor[HTML]{FCC57C}78.6 & \cellcolor[HTML]{F3E884}38.2 & \cellcolor[HTML]{FCC07B}36.5 & \cellcolor[HTML]{F8E984}37.8 & \cellcolor[HTML]{FCC37C}57.6 & \cellcolor[HTML]{DFE182}40.4 \\
MiT-b2 & \cellcolor[HTML]{63BE7B}81.2 & \cellcolor[HTML]{F8696B}18.7 & \cellcolor[HTML]{C6DB81}40.5 & \cellcolor[HTML]{F8696B}23.9 & \cellcolor[HTML]{B1D580}60.9 & \cellcolor[HTML]{F8696B}62.5 \\
NAT-T & \cellcolor[HTML]{FFEB84}80.1 & \cellcolor[HTML]{FFEB84}37.1 & \cellcolor[HTML]{FFEB84}38.5 & \cellcolor[HTML]{FFEB84}37.2 & \cellcolor[HTML]{FFEB84}59.3 & \cellcolor[HTML]{FFEB84}42.9 \\
 \bottomrule
\end{tabular}%
}
\end{table}

\begin{table}
\caption{Results in the Class-Incremental Setting with different normalization layers.}
\label{tab:result_ci_norm}
\resizebox{\columnwidth}{!}{%
\begin{tabular}{@{}l|c|ccc|cc}
\toprule
\multicolumn{1}{c|}{} & \textbf{Task 1} & \multicolumn{3}{c|}{\textbf{Task 2}} & \multicolumn{1}{l}{\textit{learning}} & \multicolumn{1}{l}{} \\
 
\multicolumn{1}{c|}{\multirow{-2}{*}{\textbf{Normalization Layer}}} & \textit{0-15} & \textit{0-15} & \textit{16-20} & \textit{all} & \textit{accuracy} & \textit{forgetting} \\ \midrule
Batch Norm (BN) & \cellcolor[HTML]{63BE7B}74.9 & \cellcolor[HTML]{63BE7B}23.5 & \cellcolor[HTML]{63BE7B}32.4 & \cellcolor[HTML]{63BE7B}25.7 & \cellcolor[HTML]{63BE7B}53.7 & \cellcolor[HTML]{FED781}51.4 \\
Continual Norm (CN) & \cellcolor[HTML]{9CCF7F}72.8 & \cellcolor[HTML]{E7E483}19.5 & \cellcolor[HTML]{DFE283}29.5 & \cellcolor[HTML]{E6E483}21.9 & \cellcolor[HTML]{B6D680}51.2 & \cellcolor[HTML]{FCA878}53.3 \\
Batch Re-Norm (BRN) & \cellcolor[HTML]{6CC17C}74.6 & \cellcolor[HTML]{FEEB84}18.8 & \cellcolor[HTML]{68C07C}32.3 & \cellcolor[HTML]{E2E383}22 & \cellcolor[HTML]{6AC07C}53.5 & \cellcolor[HTML]{F8696B}55.8 \\
Group Norm (GN) & \cellcolor[HTML]{F8696B}62 & \cellcolor[HTML]{F8696B}12.2 & \cellcolor[HTML]{F8696B}23.6 & \cellcolor[HTML]{F8696B}14.9 & \cellcolor[HTML]{F8696B}42.8 & \cellcolor[HTML]{E3E382}49.8 \\
Instance Norm (IN) & \cellcolor[HTML]{FA9B74}64.8 & \cellcolor[HTML]{FEEA83}18.7 & \cellcolor[HTML]{FAA075}25.8 & \cellcolor[HTML]{FEDB81}20.4 & \cellcolor[HTML]{FA9D75}45.3 & \cellcolor[HTML]{63BE7B}46.1 \\
Layer Norm (LN) & \cellcolor[HTML]{FBA877}65.5 & \cellcolor[HTML]{FCB679}16.1 & \cellcolor[HTML]{FDD880}28 & \cellcolor[HTML]{FCBC7A}18.9 & \cellcolor[HTML]{FCBC7B}46.8 & \cellcolor[HTML]{D5DF81}49.4 \\ \bottomrule
\end{tabular}%
}
\end{table}

\begin{figure}
\centering
\includegraphics[width=\columnwidth]{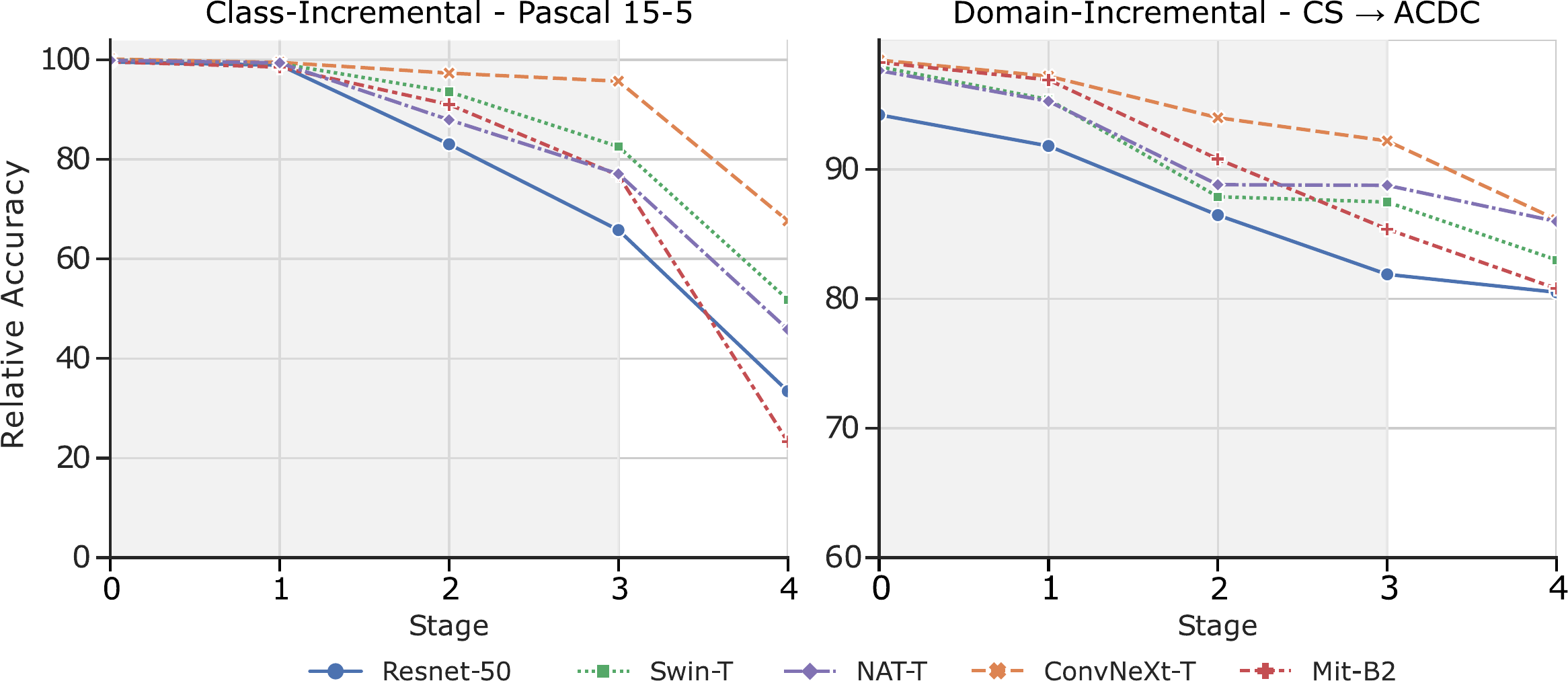}
\caption{Activation drift from $f_0$ to $f_1$ measured by relative mIoU on the first task of the Frankenstein Networks stitched together at specific stages (horizontal axis). The stages of the encoder are layer 0--3 (grey area), the decoder layers is 4 (white area). In class-incremental only the late stages of the backbone and the decoder are affected by activation drift, whereas in domain-incremental learning the activation drift already occurs at early stages. We observe that the activations of ConvNeXt-T are the most similar.}
\label{fig:layer_stitch_results}
\end{figure}

\subsection{Analyzing the Activation Drift}\label{sec:activation_drif}
We use the Dr. Frankenstein toolset~\cite{Csiszarik2021} to measure the activation drift for each stage depicted in \cref{fig:layer_stitch_results} between the model before and after learning the final task, to investigate which stages of the different networks are most affected by internal activation drift.
We follow the setup without an additional stitching layer \cite{Kalb2022}, meaning that the activations of stage $n$ under examination $f_1$ are directly propagated to the stage $n+1$ in $f_0$. 
The stitching results for the class-incremental and domain-incremental experiments are shown in \cref{fig:layer_stitch_results}. 
We see that during class-incremental learning the encoder stages up to and including stage 1 are barely affected by activation drift and only affecting the later layers, especially of the decoder. 
However, in the domain-incremental learning setting, we see that the first layers are primarily affected by activation drift and the representation shift is much less severe than in the class-incremental setting.
These observations also suggest that in class-incremental learning, the choice of the decoder could play a much more important role than in the domain-incremental setting, as activation drift is mainly affecting the decoder layers.

\subsection{Effect of the Decoder Architecture}\label{sec:decoderhead}
Previous experiments have shown that in the class-incremental setting, forgetting occurs primarily in the decoder stages of a model.
Therefore, we investigate how different decoder architectures affect the performance in continual learning using the same ConvNeXt-T backbone. 
We note that we use multi-stage input for UperNet, FPN and SegFormer-Head as proposed in their respective implementations and only the single-stage output (last stage of the encoder) for ASPP, FCN and PSPNet. 
The results are displayed in \cref{tab:result_ci_dec}. 
We observe that the decoder heads using multi-stage input from the decoder achieve higher learning accuracy, but suffer more from forgetting. 
We assume that the multi-stage input leads to changes in earlier stages so that early stages are already affected by activation drift. 
Furthermore, we find that, despite using fewer parameters, ASPP outperforms UperNet-Head on all classes. 
Finally, the Segformer-Head achieves comparable performance to the UperNet-Head while having 1/20\textsuperscript{th} of its parameters.

\section{Ablation on ConvNext}\label{sec:convnext} 
\begin{figure}
	\centering
     \fontsize{7pt}{9pt}
        \def\svgwidth{1.001\columnwidth}
	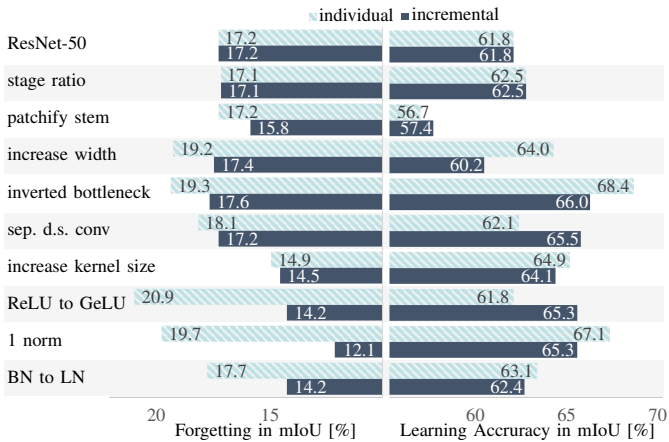
	\caption{Forgetting and learning accuracy for ResNet $\rightarrow$ ConvNeXt individual vs. incremental changes. Striped light blue bars indicate changes that are made individually on ResNet, while solid blue bars include all previous changes. The sequence of changes is from top to bottom.}
	\label{fig:convnext_abl_com}
\end{figure}

Since ConvNeXt is the best-performing architecture in our experiments, we ablate architectural changes starting from ResNet-50 in the context of continual learning to see the contribution of each change individually and incrementally without reverting the previous changes, similar to Liu \etal~\cite{Convnext}.    
We make these changes to the ResNet-50 architecture as described in \cref{tab:convnext_changes} and train it in the domain-incremental setting.
For a fair comparison, we use a model without pre-training and use the same random initialization for our experiments. 
The \textit{learning accuracy} and \textit{forgetting} for the individual and incremental changes are reported in \cref{fig:convnext_abl_com}.
Some individual changes like increasing the width from 64 to 96, inverted bottleneck and reducing the number of normalization layers to 1 (1norm) improve learning accuracy, but in turn also increase forgetting.
Increasing the kernel size is the only individual change that increases \textit{learning accuracy} and decreases \textit{forgetting}.
However, when the changes are made incrementally, they lead to a reduction in forgetting and a slight improvement in learning accuracy.
One of the interesting findings is that if we reduce the normalization layers from 3 to 1 (1norm) it leads to an increase in forgetting by +2.5 mIoU, but when done incrementally, it leads to an improvement in forgetting of -2.1 mIoU.
For the incremental changes, the most effective additions were: \textit{patchify stem}, \textit{increasing kernel size} and \textit{1 norm}. 
These changes coincide with the architectural changes Wang \etal \cite{robCNN2} proposed to make CNNs as robust to out-of-distribution samples as VTs. 
This further strengthens our claim that the ability of the model to mitigate forgetting depends on its ability to learn general and robust features.

\begin{table}
\renewcommand*{\arraystretch}{1.08}
\caption{Architectural changes in our ablation ResNet-50 $\rightarrow$ ConvNeXt inspired by Liu \etal~\cite{Convnext}.}
\label{tab:convnext_changes}
\resizebox{\columnwidth}{!}{%
\begin{tabular}{cll}
\toprule
\textbf{No.} & \textbf{Name} & \textbf{Description} \\ \midrule
1 & stage ratio & \begin{tabular}[c]{@{}l@{}}change the number of blocks in each stage from $(3, 4, 6, 3)$ \\ to $(3, 3, 9, 3)$\end{tabular} \\ \rowcolor[HTML]{F5F5F5}
2 & patchify stem & \begin{tabular}[c]{@{}l@{}}stem cell was replaced with a patchify layer with \\ $4\times4$ stride $4$ Convolutional Layers\end{tabular} \\ 
3 & increase width & Change width [64 to 96] \\ \rowcolor[HTML]{F5F5F5}
4 & inverted bottleneck & \begin{tabular}[c]{@{}l@{}}Change the channels in the block from $(64,64,256)$ \\ to $(96,384,96)$\end{tabular} \\
5 & sep. d.s. conv & \begin{tabular}[c]{@{}l@{}}Use depth-wise separable convolutions to reduce \\ computation as a trade-off for using large kernels\end{tabular} \\ \rowcolor[HTML]{F5F5F5}
6 & increase kernel size & Change Kernel size from $[3\times3]$ to $[7\times7]$ \\
7 & ReLU to GeLU & Change the activation layer from ReLU to GeLU \\ \rowcolor[HTML]{F5F5F5}
8 & 1 norm & \begin{tabular}[c]{@{}l@{}}Reduce the number of normalisation layers by removing\\  two out of three batch normalization layers\end{tabular} \\ 
9 & BN to LN & \begin{tabular}[c]{@{}l@{}}Change the normalisation layer form batch normalisation \\ to layer normalization\end{tabular} \\ \bottomrule
\end{tabular}%
}

\end{table}

\begin{table}
\caption{Results in the Class-Incremental Setting with different decoder architectures on top of ConvNeXt backbone.}
\label{tab:result_ci_dec}
\resizebox{\columnwidth}{!}{%
\begin{tabular}{@{}lccccccc}
\toprule
\multicolumn{1}{c}{} & \textbf{Task   1} & \multicolumn{3}{c}{\textbf{Task 2}} & \multicolumn{1}{l}{\textit{learning}} & \multicolumn{1}{l}{} & \multicolumn{1}{l}{} \\ \cmidrule(l){2-8} 
\multicolumn{1}{c}{\multirow{-2}{*}{\textbf{Decoder}}} & \textit{0-15} & \textit{0-15} & \textit{16-20} & \textit{all} & \textit{accuracy} & \textit{forgetting} & \#params \\ \midrule
UperNet-Head & \cellcolor[HTML]{63BE7B}81.1 & \cellcolor[HTML]{DEE283}50.4 & \cellcolor[HTML]{72C37C}43.9 & \cellcolor[HTML]{D3DF82}48.9 & \cellcolor[HTML]{63BE7B}62.5 & \cellcolor[HTML]{FFE683}30.7 & \cellcolor[HTML]{F8696B}31.4M \\
FPN-Head & \cellcolor[HTML]{A0D07F}79.5 & \cellcolor[HTML]{F8696B}34.1 & \cellcolor[HTML]{CCDD82}43.3 & \cellcolor[HTML]{F8696B}36.3 & \cellcolor[HTML]{AAD380}61.4 & \cellcolor[HTML]{F8696B}45.4 & \cellcolor[HTML]{63BE7B}1.4M \\
SegFormer-Head & \cellcolor[HTML]{BFD981}78.7 & \cellcolor[HTML]{FEE582}49.1 & \cellcolor[HTML]{F97F6F}42.0 & \cellcolor[HTML]{FEE282}47.4 & \cellcolor[HTML]{E9E583}60.4 & \cellcolor[HTML]{F6E883}29.6 & \cellcolor[HTML]{68BF7B}1.8M \\
FCN-Head & \cellcolor[HTML]{F8696B}74.8 & \cellcolor[HTML]{FCBE7B}44.4 & \cellcolor[HTML]{FCC37C}42.6 & \cellcolor[HTML]{FCBD7B}44.0 & \cellcolor[HTML]{F8796E}58.7 & \cellcolor[HTML]{FFE884}30.4 & \cellcolor[HTML]{ECE582}11.8M \\
PSPNet-Head & \cellcolor[HTML]{F9806F}75.2 & \cellcolor[HTML]{7DC67D}52.3 & \cellcolor[HTML]{F8696B}41.8 & \cellcolor[HTML]{9ECF7F}49.8 & \cellcolor[HTML]{F8696B}58.5 & \cellcolor[HTML]{6BC07B}22.9 & \cellcolor[HTML]{FFE283}14.6M \\
ASPP & \cellcolor[HTML]{F98670}75.3 & \cellcolor[HTML]{63BE7B}52.8 & \cellcolor[HTML]{63BE7B}44.0 & \cellcolor[HTML]{63BE7B}50.8 & \cellcolor[HTML]{FDCD7E}59.7 & \cellcolor[HTML]{63BE7B}22.5 & \cellcolor[HTML]{FCA477}23.2M \\
\bottomrule
\end{tabular}%
}
\end{table}

\section{CONCLUSIONS}
We conducted experiments in domain- and class-incremental learning with a variety of CNNs, VTs, and hybrid architectures, as well as different normalization layers, to investigate the effects of architecture choices in continual semantic segmentation.
We found that the choice of architecture has a significant impact on performance in terms of learning accuracy and mitigating the effects of catastrophic forgetting. 
Specifically, we found that the traditional CNNs show higher plasticity than their VT counterparts, but in turn are more susceptible to forgetting.
Hybrid architectures combine the flexibility of CNNs with the stability of VTs by adding the inductive bias of CNN architectures to the VTs.
Even in class-incremental learning, where forgetting occurs primarily in decoder layers, the choice of backbone architecture has a significant impact on forgetting.
Furthermore, the experiments show that input into the decoder from multiple stages of the backbone improves learning accuracy but increases forgetting because early layers are more easily affected by activation drift.
Additionally, our experiments conclude that Continual Norm has the best trade-off in terms of stability and plasticity in continual semantic segmentation.
Overall, we find ConvNeXt performs the best in our experiments.
To better understand how the ResNet block benefited from the architectural changes introduced by ConvNeXt, we conducted a second ablation study, which revealed that, with the exception of a larger kernel size, none of the proposed changes individually resulted in better performance.
Only when they are used in combination, we find that specifically patchifying the stem, increasing the kernel size and using fewer normalization layers lead to a significant reduction in forgetting.
Overall, our work demonstrates the importance of architecture in the development of algorithms for continual learning and lays the groundwork for future research into architectures tailored to continual semantic segmentation.

\section*{ACKNOWLEDGMENT}
The research leading to these results is funded by the German Federal Ministry for Economic Affairs and Climate Action within the project “KI Delta Learning“ (Förderkennzeichen 19A19013T).

{\small
\bibliographystyle{IEEEtran}
\bibliography{IEEEabrv,references}
}

\end{document}